\documentclass[sn-mathphys-num]{sn-jnl}

\usepackage{graphicx}%
\usepackage{multirow}%
\usepackage{amsmath,amssymb,amsfonts}%
\usepackage{amsthm}%
\usepackage{mathrsfs}%
\usepackage[title]{appendix}%
\usepackage{xcolor}%
\usepackage{textcomp}%
\usepackage{manyfoot}%
\usepackage{booktabs}%
\usepackage{algorithm}%
\usepackage{algorithmicx}%
\usepackage{algpseudocode}%
\usepackage{listings}%
\usepackage{soul,color}
\usepackage[style=authoryear, sorting=nyt, maxcitenames=1, mincitenames=1, maxbibnames=99, minbibnames=99]{biblatex} 
\DeclareFieldFormat[article]{title}{#1} 
\DeclareFieldFormat[book]{title}{#1}    
\DeclareFieldFormat[incollection]{title}{#1} 
\DeclareFieldFormat[inproceedings]{title}{#1} 
\DeclareFieldFormat[techreport]{title}{#1} 

\renewbibmacro{in:}{}

\DeclareFieldFormat[article]{pages}{#1} 
\DeclareFieldFormat[book]{pages}{#1}    

\DeclareFieldFormat{doi}{\url{#1}} 

\theoremstyle{thmstyleone}%
\theoremstyle{thmstyletwo}%

\theoremstyle{thmstylethree}%

\begin{document}

\title[General Information Metrics for Improving AI Model Training Efficiency]{General Information Metrics for Improving AI Model Training Efficiency}


\author*[1]{Jianfeng Xu}\email{xujf@sjtu.edu.cn}
\equalcont{These authors contributed equally to this work.}
\author*[1]{Congcong Liu}\email{sjtu6418516@sjtu.edu.cn}
\author[3]{Xiaoying Tan}
\equalcont{These authors contributed equally to this work.}
\author[4]{Xiaojie Zhu}
\equalcont{These authors contributed equally to this work.}
\author[2]{Anpeng Wu}
\equalcont{These authors contributed equally to this work.}
\author[4]{Huan Wan}
\author[4]{Weijun Kong}
\author[1]{Chun Li}
\author[1]{Hu Xu}
\author[2]{Kun Kuang}
\author*[2]{Fei Wu}\email{wufei@zju.edu.cn}

\affil[1]{\orgdiv{Institute for Smart Courts}, \orgname{Shanghai Jiao Tong University}, \orgaddress{\city{Shanghai}, \postcode{200030}, \country{China}}}
\affil[2]{\orgdiv{Department of Computer Science and Technology}, \orgname{Zhejiang University}, \orgaddress{\city{Hangzhou}, \postcode{310027}, \country{China}}}
\affil[3]{\orgdiv{China Judicial Big Data Research Institute Co., Ltd.}, \orgaddress{\city{Beijing}, \postcode{100035}, \country{China}}}
\affil[4]{\orgdiv{iFLYTEK Co., Ltd.}, \orgaddress{\city{Hefei}, \postcode{230088}, \country{China}}}






\abstract{To address the growing size of AI model training data and the lack of a universal data selection methodology—factors that significantly drive up training costs—this paper presents the General Information Metrics Evaluation (GIME) method. GIME leverages general information metrics from Objective Information Theory (OIT), including \textit{volume}, \textit{delay}, \textit{scope}, \textit{granularity}, \textit{variety}, \textit{duration}, \textit{sampling rate}, \textit{aggregation}, \textit{coverage}, \textit{distortion}, and \textit{mismatch} to optimize dataset selection for training purposes. Comprehensive experiments conducted across diverse domains, such as CTR Prediction, Civil Case Prediction, and Weather Forecasting, demonstrate that GIME effectively preserves model performance while substantially reducing both training time and costs. Additionally, applying GIME within the Judicial AI Program led to a remarkable 39.56\% reduction in total model training expenses, underscoring its potential to support efficient and sustainable AI development.}




\maketitle
\section{Introduction}

Artificial intelligence (AI) is transforming numerous aspects of contemporary life, with advancements fueled largely by the training of models on extensive datasets~\parencite{pouyanfar2018survey,dong2021survey,bialkova2024ai}. This is particularly evident in areas like autonomous driving~\parencite{liu2024afm3d,cui2024survey}, generative AI \parencite{feuerriegel2024generative,huang2024federated}, and medical image processing \parencite{tian2024role,alzubaidi2024ssp}, which depend on large-scale model training. As these models expand to encompass hundreds of billions of parameters, the need for high-quality training data becomes critical~\parencite{zhao2023survey,minaee2024large}. Training such large-scale models often requires tens to hundreds of trillions of tokens, substantial interdisciplinary effort over months, and a vast array of computational resources, including thousands of GPUs and high levels of energy consumption~\parencite{achiam2023gpt,touvron2023llama,touvron2023llama2,chowdhery2023palm}. A core challenge is ensuring that training data is meticulously curated—ineffective data selection can yield models that underperform, fall short of desired objectives, and waste considerable resources~\parencite{chowdhery2023palm,gunasekar2023textbooks}. Thus, once model architecture and algorithms are defined, the quality of the training data becomes paramount to a model’s success, significantly influencing the performance and relevance of AI technologies across various domains~\parencite{hamid2023data,zha2023data}.By focusing on data quality, small-scale models can achieve performance comparable to much larger models. For instance, Phi-1.5 achieves performance on par with models 5 times its size, while Phi-2 matches or even surpasses the performance of models 25 times larger\parencite{gunasekar2023textbooksneed, li2023textbooksneediiphi15}.

\begin{figure}[H]
 \centering
 \includegraphics[width=0.6\textwidth]{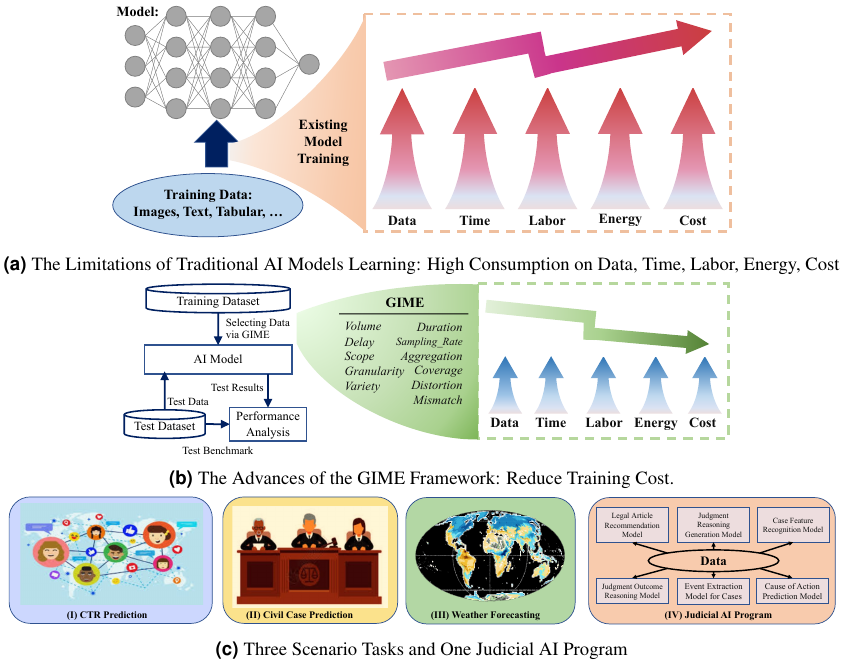}
 \caption{Overview of General Information Metrics Supporting AI Model Training Data Selection. (a) Traditional AI model training often lacks effective methods for selecting optimal data, leading to significant resource costs. (b) By employing 11 general information metrics in the GIME algorithm, i.e., \textit{volume}, \textit{delay}, \textit{scope}, \textit{granularity}, \textit{variety}, \textit{duration}, \textit{sampling rate}, \textit{aggregation}, \textit{coverage}, \textit{distortion}, and \textit{mismatch}, we can select AI training datasets and substantially reduce resource consumption during the training process. (c): This paper presents experiments that apply general information metrics to support AI model training in three scenario tasks and judicial program: (I) Click-Through Rate (CTR) Prediction, predicting the likelihood of a user clicking on a specific item, a key task in online advertising and recommendation systems. (II) Civil Case Prediction, using civil case data to forecast case types, supporting judges, lawyers, litigants, and researchers. (III) Weather Forecasting, predicting future weather conditions using meteorological data from multiple cities, a long-standing AI application. (IV) Judicial AI Program in training models for the cause of action prediction, case feature recognition, event extraction, judgment outcome reasoning, legal article recommendation, and judgment reasoning generation.}
 \label{fig:1}
\end{figure}

Current approaches to training data selection encompass full data utilization, random sampling, stratified sampling, information-theoretic feature selection, data augmentation, active learning, and transfer learning~\parencite{motamedi2021data,zha2023data,singh2023systematic,jakubik2024data}. While full data utilization offers comprehensiveness, it is also resource-intensive and frequently inefficient, as it does not guarantee performance gains. Random sampling, a widely used and straightforward technique, can produce inconsistent results and lead to excessive costs. Stratified sampling relies on specific data structures, whereas information-theoretic feature selection depends on prior statistical assumptions, which limits its adaptability. Data augmentation~\parencite{shorten2019survey,motamedi2021data} can enhance datasets that lack variety or density, but its applicability across different domains is often constrained. Active learning~\parencite{danka2018modal,zhan2022comparative} and transfer learning prioritize data based on model-specific objectives, inherently tying them to particular models and reducing their generalizability. Despite widespread reliance on criteria such as completeness, heterogeneity, variety, accuracy, timeliness, and balance to assess training data, these metrics remain largely subjective, lacking strong theoretical foundations or standardized guidelines. Consequently, the absence of a universally effective data selection method—one adaptable across diverse domains and models—constitutes a significant gap in AI development~\parencite{gebru2021datasheets}. Addressing this issue is crucial for advancing AI in a manner that is both sustainable and efficient.

The physical world is fundamentally composed of matter, energy, and information~\parencite{program1976annual,shapiro1999information,wiener2019cybernetics}.
Scientific research on matter and energy has always relied on measurable indicators such as volume, weight, temperature, heat, power, and so on. Only through quantifiable metrics can universal mathematical expressions be formulated, revealing underlying principles through quantitative laws~\parencite{wiener2019cybernetics}. Training datasets used in AI function similarly, serving as critical information resources. To tackle the challenge of dataset selection, it is essential to first establish a set of clear metrics to evaluate these information resources, which can then guide assessments of a dataset's suitability for supporting AI model training~\parencite{kaplan2020scaling}. Shannon’s information theory, which introduced information entropy as a means of assessing data, has proven useful in training dataset selection. However, calculating information entropy requires prior knowledge of event probability distributions—an often impractical requirement in real-world scenarios~\parencite{renyi1961measures,ash2012information}. This limitation makes it challenging to generalize information entropy as a tool for AI model training~\parencite{sekeroglu2022comparative,bandi2023power}.

Building on this perspective, Xu et al., introduced Objective Information Theory (OIT)~\parencite{xu2014objective,xu2023objective}, which defines 11 general information metrics~\parencite{xu2022foundations}: \textit{volume}, \textit{delay}, \textit{scope}, \textit{granularity}, \textit{variety}, \textit{duration}, \textit{sampling rate}, \textit{aggregation}, \textit{coverage}, \textit{distortion}, and \textit{mismatch}. These metrics offer a comprehensive framework that integrates key principles from information science, providing a solid theoretical basis for evaluating AI training datasets~\parencite{xu2024research}. By evaluating datasets with these general information metrics before training begins, researchers can better anticipate the effectiveness of the training process. This systematic assessment enhances training efficiency and reduces the likelihood of wasting valuable resources—both time and computational power—on data that may be suboptimal or misaligned with model objectives. 

In this work, based on the mathematical definitions of 11 general information metrics, we designed tailored methods to calculate these metrics for AI model training tasks in three different domains: CTR Prediction~\parencite{guo2017deepfm}, Civil Case Prediction~\parencite{ma2021legal}, and Weather Forecasting~\parencite{abhishek2012weather}. Through extensive experimentation, we analyzed the correlation between these metrics and model performance, finding that changes—whether increases or decreases—in most metrics are directly linked to improvements in model outcomes. Building on these insights, we introduced the General Information Metrics Evaluation (GIME) method, which integrates a data evaluation phase before training begins. This pre-training assessment phase ensures that the training process only proceeds if relevant dataset metrics meet established thresholds, thereby reducing the likelihood of inefficient or unsuccessful training sessions. GIME significantly optimizes the training process, lowering training costs while preserving model performance. Our experimental validation across the three tasks demonstrated that GIME can effectively reduce training time and enhance efficiency without compromising model quality. Furthermore, we systematically applied the GIME framework within AI-driven solutions in the judicial program, yielding significant reductions in data size, human labor, energy consumption, and development costs. 

The GIME method, based on the General Information Metric proposed by OTI, offers several unique advantages over existing training data selection methods: (1) it is universally applicable across domains, (2) independent of specific AI model structures or algorithms, (3) not constrained by data scale or structure, (4) requires no prior knowledge of data distribution, and (5) eliminates the need for labor-intensive preprocessing such as labeling or augmentation. Therefore, it can serve as a universally applicable data selection method for model training, making AI model outcomes more predictable and significantly reducing costs.

\section{Methodology}

\subsection{Common Methods for AI Model Training Data Selection}
In the pursuit of effective data selection and utilization strategies, a commonly employed baseline is to use the entire dataset for training (full data utilization). Although this approach may capture comprehensive information, it often proves highly inefficient in large-scale, diverse datasets. The computational and resource costs are substantial, and such exhaustive usage does not necessarily guarantee performance improvements \parencite{motamedi2021data,jakubik2024data}. In response, various subset selection techniques have emerged.

Random sampling is one of the simplest strategies, selecting subsets without any specific criteria, such as \parencite{okanovic2023repeated}, \parencite{wongvorachan2023comparison}. Despite its ease of implementation, it often fails to represent underlying data structures, leading to unstable model performance and inefficient use of resources \parencite{singh2023systematic,jakubik2024data}. A more structured approach is stratified sampling, such as \parencite{jiao2022hierarchical}, \parencite{wang2019efficient} .which divides the dataset into non-overlapping strata and selects samples proportionally, thereby preserving key data distributions. However, this method relies on prior knowledge of the underlying data distribution, limiting its applicability in scenarios where such information is unavailable \parencite{singh2023systematic,jakubik2024data}.

Information-theoretic feature selection methods, such as \parencite{wan2022r2ci}, \parencite{pawluk2019information}, seek to enhance model performance by maximizing information gain and minimizing redundancy, employing measures like entropy and mutual information. Yet, these approaches often require accurate probability distributions—a non-trivial demand in complex, real-world contexts—thus restricting their general scalability \parencite{renyi1961measures,ash2012information}. Data augmentation, by contrast, increases training data diversity and density through transformations of existing samples, such as \parencite{cubuk2019autoaugment}, \parencite{borovykh2017conditional}. Although successful in certain domains (e.g., computer vision), its efficacy and adaptability are limited across varied data types and tasks \parencite{motamedi2021data,shorten2019survey}.

Active learning, such as modAL\parencite{danka2018modal}, \parencite{yoo2019learning}, \parencite{park2020robust}, adopts an iterative procedure, starting with a small labeled dataset and incrementally selecting the most informative samples for annotation based on model-specific uncertainty criteria. While it can reduce labeling costs and improve model performance, its effectiveness heavily depends on the chosen model and uncertainty measures, thereby constraining its broader applicability \parencite{danka2018modal,zhan2022comparative}.  Transfer learning such as \parencite{cai2016batch}, \parencite{finn2017model}, which leverages models pretrained on large datasets, can expedite performance gains on smaller target datasets. However, its benefits wane when the target domain markedly differs from the source domain, thereby limiting its universality \parencite{singh2023systematic,jakubik2024data}.

In summary, current methods for AI training data selection face limitations such as computational inefficiency, reliance on specific data characteristics, and lack of generalizability. These challenges highlight the need for adaptive methods that balance efficiency, accuracy, and scalability across diverse tasks. The general information metrics constructed by OIT offer a potential solution to address these questions.

\subsection{Definition and Implications of the General Information Metrics}

Objective Information Theory (OIT) defines information as follows: Let $\mathbb{O}$ denote the set of all content of the objective world, let $\mathbb{S}$ denote the set of content of the subjective world, and let $\mathbb{T}$ be a set of times. The elements in these sets can be specified according to the requirements in different areas. A noumenon o is thus an element of the power set $2^{\mathbb{O}\cup\mathbb{S}}$ (or a subset of $\mathbb{O}\cup\mathbb{S}$), the occurrence duration is $T_h\in2^\mathbb{T}$ , and $f\left(o,T_h\right)$ is the set of states of o over $T_h$. The carrier $c\in2^\mathbb{O}$, the set of reflection times $T_m \in 2^\mathbb{T}$, and the set of reflection states $g\left(c,T_m\right)$ are all nonempty sets. Information $I$ is thus an enabling mapping from $f\left(o,T_h\right)$ to $g\left(c,T_m\right)$, which is expressed as $I:f\left(o,T_h\right)\rightarrow g\left(c,T_m\right)$ or $I\left(f\left(o,T_h\right)\right)=g\left(c,T_m\right)$. The collection of all information is referred to as the information space, denoted by $\mathfrak{T}$, which is one of the three essential elements of the objective world according to OIT.

Information can also be represented as a sextuple, i.e., $I = \langle o, T_h, f, c, T_m, g \rangle$. When $I$ is an injective mapping from $f(o, T_h)$ to $g(c, T_m)$, meaning there exists an inverse mapping $I^{-1}$ such that $I^{-1}(g(c, T_m)) = f(o, T_h)$, it is called restorable information. For restorable information, Xu et al. \parencite{xu2023objective,xu2022foundations,xu2024research} defines 11 general information metrics as shown in Table \ref{tab:OIT}.

\begin{table}[h]
\caption{\centering 11 General Information Metrics}\label{tab:OIT}%
\begin{tabular}{c}
\includegraphics[width=0.85\textwidth]{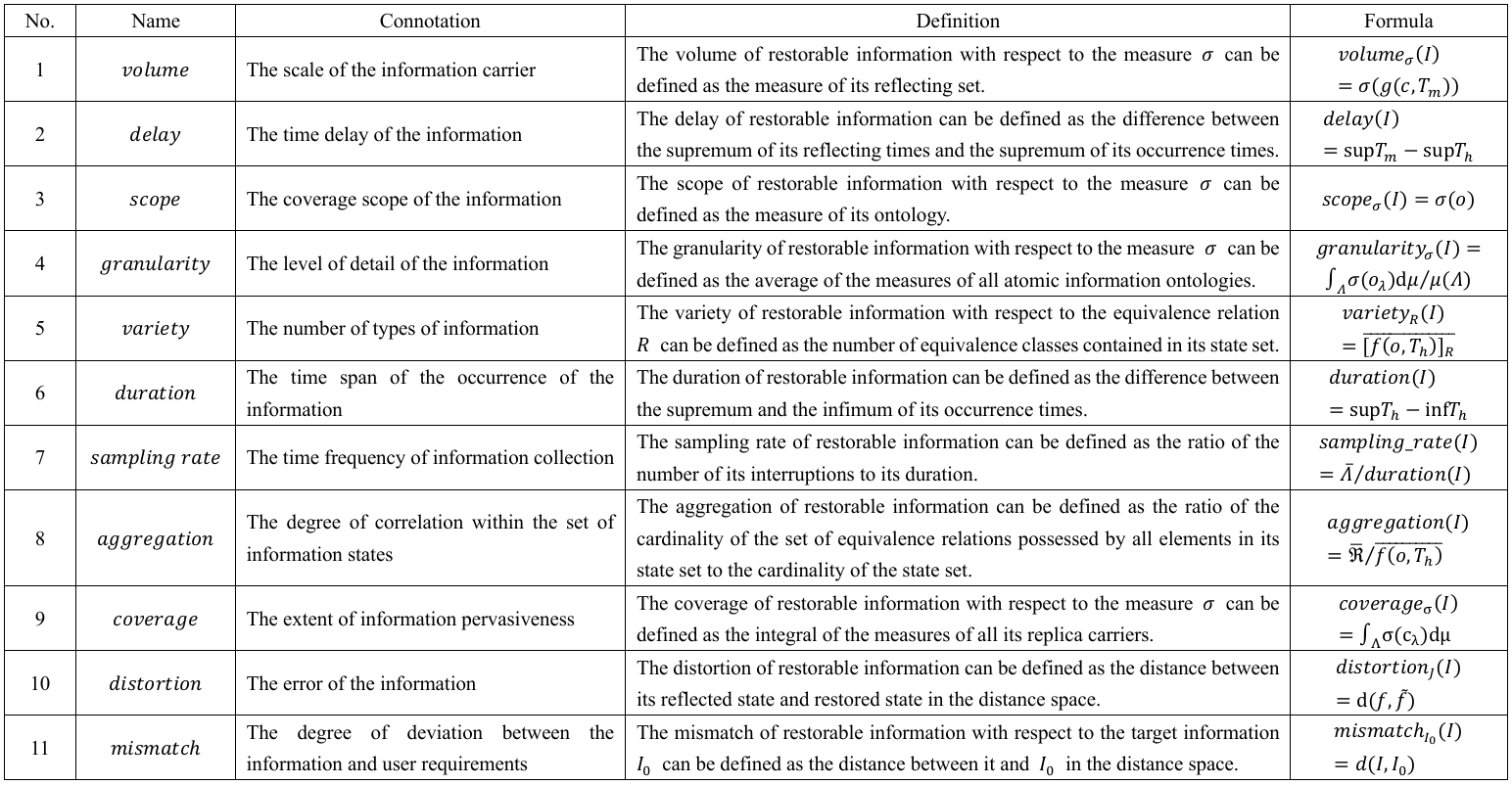} \\
\end{tabular}
\end{table}

According to the perspective of OIT, data is a manifestation of information. Based on the definitions and formulas in Table 1, the 11 general information metrics are independent of data type, scale, and structure, making them applicable to the evaluation of datasets across all domains. Among these metrics, \textit{volume} is correlated with \textit{scope}, \textit{granularity}, \textit{variety}, \textit{duration}, and \textit{sampling rate}. It is not difficult to demonstrate that under certain typical conditions, such as video information, the following relationship holds:
\begin{equation}
volume=k \cdot \frac{{ scope } \cdot { variety } \cdot { duration } \cdot { sampling\_rate }}{{ granularity }}
\end{equation}
where $k$ is a proportional factor. Therefore, in research, it is possible to adjust \textit{volume} by manipulating metrics such as \textit{scope}, \textit{granularity}, \textit{variety}, \textit{duration}, and \textit{sampling rate}. Furthermore, it can be observed that the remaining 10 metrics are independent of one another, with each uniquely characterizing an important property of the dataset. Consequently, without the need to predefine the event probability distribution, this approach eliminates the previous limitation of relying solely on \textit{volume} as the universal metric for data evaluation. Instead, up to 11 metrics, including 10 fully independent ones, can now be utilized to assess datasets. This undoubtedly greatly enhances the comprehensiveness, precision, and generality of data evaluation.



\subsection{GIME Method for AI Model Training Data Selection}

\begin{figure}[t]
 \centering
 \includegraphics[width=0.8\textwidth]{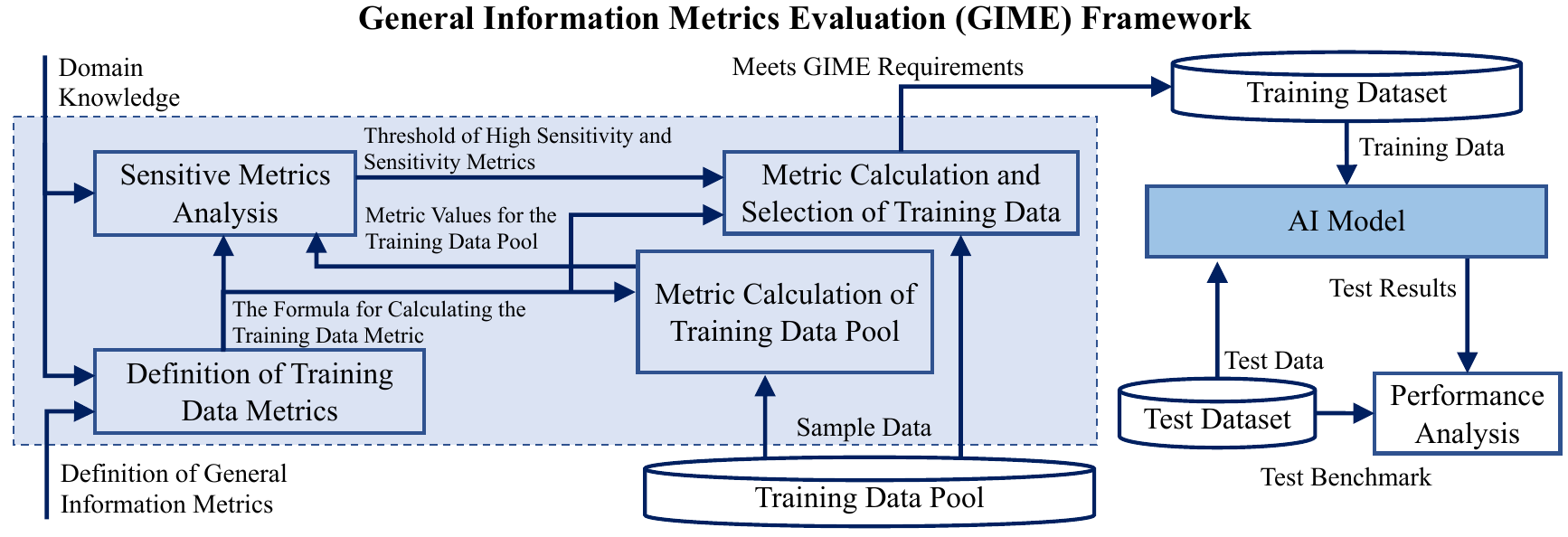}
 \caption{Work Pipeline of General Information Metrics Evaluation (GIME) Framework. The light gray shaded area represents the GIME framework, which comprises four modules. GIME selects data from the training data pool, and once the information metrics of the chosen dataset meet the threshold criteria, the AI model training process is initiated. Subsequently, the model’s performance is evaluated and analyzed using a test dataset. }
 \label{fig:method}
\end{figure}

In this paper, we propose a novel GIME method, which integrates a data selection and evaluation process into a standard AI model training framework (highlighted in gray), as shown in Figure \ref{fig:method}. This process is independent of the application domain and model type. The “training data pool” consists of all available relevant data, while the “training dataset” is a subset selected from the pool after evaluation by GIME to meet specified criteria.

The GIME framework comprises four key modules, as shown in Figure \ref{fig:method}. In the "Definition of Training Data Metrics" module, we define the calculation formulas for training data metrics, based on general information metrics and domain-specific knowledge, ensuring metric values can be derived for any dataset. Then, in the "Metric Calculation of Training Data Pool" module, we calculate these metrics for the entire data pool, providing essential insights for setting threshold values. The "Sensitivity Metrics Analysis" module classifies metrics into high sensitivity, moderate sensitivity, and low sensitivity categories. High-sensitive metrics aim for optimal values, moderate-sensitive metrics aim for reasonable values, while low-sensitive metrics are not prioritized. These thresholds guide data selection. The "Metric Calculation and Selection of Training Data" module calculates 11 general information metrics for sampled data from the training data pool. If all thresholds are met, the data becomes the training dataset. If not, the process continues with new data until all requirements are satisfied, ensuring the optimal dataset for model training.

Among the four modules described above, only the second and fourth modules involve computational tasks, which are focused on calculating dataset metrics. Based on the definitions and formulas, once the dataset is determined, its various metrics can be obtained through simple calculations. For particularly large datasets, sampling methods can be employed to significantly reduce computational costs while estimating these metrics. Consequently, compared to the computational overhead required for model training, the additional costs introduced by the GIME method for metric calculation and evaluation are nearly negligible.


\subsection{Theoretical Proof of GIME’s Superiority}
The GIME method selects a subset of the full dataset for model training based on dataset metric thresholds, significantly reducing training costs compared to method of full data utilization. Another comparable approach is random sampling. Furthermore, the theoretical advantages of the GIME method can also be clearly demonstrated.

\noindent \textbf{Lemma:} Let the cardinality of a finite dataset $D$ be $| D |$, and $\textit{metr}$ be a metric defined on $D$. Assume that the elements of $D$ are uniformly distributed with respect to $\textit{metr}$. Define $m = \min \{\textit{metr}(s) \mid s \subset D \}$ and $M = \max \{\textit{metr}(s) \mid s \subset D \}$. Let $R$ be a randomly selected subset of $D$ with cardinality $| R | = k | D |$, where $0 < k < 1$. When $\textit{metr}$ is an additive, maximum, minimum, or mean-type metric, the mathematical expectation of $\textit{metr}(R)$ is given by $kM$, $m + k | D | \frac{M - m}{k | D | + 1}$, $M - k | D | \frac{M - m}{k | D | + 1}$, and $\frac{M + m}{2}$.

From this, we can derive the following important conclusion.

\noindent \textbf{Theorem:} Let $\textit{metr}$ be an additive, maximum, minimum, or mean-type metric defined on a finite dataset $D$, and let the elements of $D$ be uniformly distributed with respect to $\textit{metr}$. Define $m = \min \{\textit{metr}(s) \mid s \subset D \}$ and $M = \max \{\textit{metr}(s) \mid s \subset D \}$. Assume that $s$ is a subset of $D$ used for model training, and the model performance $p(s)$ is positively correlated with the metric $\textit{metr}(s)$. Let $S$ be a subset of $D$ with $| S | = k | D |$, where $0 < k < 1$, and $\textit{metr}(S) = M$. Then, for any randomly selected subset $R$ of $D$ with $| R | = | S |$, the model trained on $S$ will have statistically better performance than the model trained on $R$.

\noindent \textbf{Proof:}
Since $R$ is a randomly selected subset of $D$ with $| R | = | S | = k | D |$, the mathematical expectation of $\textit{metr}(R)$, according to the lemma, is:
\[
\begin{aligned}
\mathbb{E}(\textit{metr}(R)) =
\begin{cases}
kM, & \text{if } \textit{metr} \text{ is additive}, \\
m + k | D | \frac{M - m}{k | D | + 1}, & \text{if } \textit{metr} \text{ is maximum-type}, \\
M - k | D | \frac{M - m}{k | D | + 1}, & \text{if } \textit{metr} \text{ is minimum-type}, \\
\frac{M + m}{2}, & \text{if } \textit{metr} \text{ is mean-type}.
\end{cases}
\end{aligned}
\]
By assumption, the model performance $p(s)$ for a training dataset $s$ is positively correlated with the metric $\textit{metr}(s)$. Thus, for any subsets $P$ and $Q$ of $D$, if $\textit{metr}(P) > \textit{metr}(Q)$, then $p(P) > p(Q)$.

Since $\textit{metr}(S) = M = \max \{\textit{metr}(s) \mid s \subset D  \}$, and comparing this to the expectation results in the lemma, we have $\textit{metr}(S) = M > \mathbb{E}(\textit{metr}(R) )$, regardless of whether $\textit{metr}$ is additive, maximum, minimum, or mean-type. This indicates that, in a statistical sense, $\textit{metr}(S) > \textit{metr}(R)$. Therefore, $p(S) > p(R)$.Thus, the theorem is proven.

A detailed analysis reveals that, under general circumstances, the defined general information metrics can be categorized as follows: $\textit{volume}$, $\textit{scope}$, $\textit{variety}$, $\textit{duration}$, $\textit{aggregation}$ and $\textit{coverage}$ are additive metrics; $\textit{delay}$ is a maximum-type metric; and $\textit{granularity}$, $\textit{sampling rate}$, $\textit{distortion}$ and $\textit{mismatch}$ are mean-type metrics.
Furthermore, regardless of whether each metric of the training dataset is positively or negatively correlated with model performance, the use of the GIME method ensures that, when the threshold of a high-sensitivity metric reaches the optimal value for the training data pool, the resulting training dataset will statistically outperform random sampling methods in terms of model performance.

\section{Test and Application Results}


\subsection{Three Test Scenarios and models}
To evaluate the effectiveness of general information metrics in improving the efficiency of AI model training, we conducted experimental validations across three entirely different domains and corresponding models. 
Specifically, we select CTR Prediction (represents Human Behavior tasks), Civil Case Prediction (represents Natural Language Understanding tasks), and Weather Forecasting (represents Time-Series tasks) to evaluate GIME’s ability to reduce data size, training time, and energy consumption in large-scale complex tasks.

In CTR Prediction, we study the impact of online advertising recommendations on user Click-Through Rates and estimate CTR via deep learning models \parencite{yang2022click}. We utilized the Avazu dataset, a large-scale open-source dataset containing approximately 40 million mobile ad click records, which is widely used in both academic and industry \parencite{song2020towards}. This dataset includes fields such as ID, click labels, timestamps, anonymous categorical variables, and various attributes related to websites, applications, and devices. Second, in Civil Case Prediction \parencite{cui2023survey}, we predict civil case types based on court judgment documents, which is a fundamental task in providing AI-assisted services to judges, lawyers, litigants, and legal researchers. We collected data from the China Judgments Online platform \parencite{shi2021smart}, the largest open-source collection of court judgments in the world, containing more than 200 million judgments. Each document includes information such as court details, case type, anonymized litigant information, case facts, applicable laws, and the final ruling. Finally, in Weather Forecasting tasks, we selected weather data from nine European cities (Basel, Budapest, De Bilt, D¨usseldorf, London, Ljubljana, Maastricht, Munich, and Oslo) available on Kaggle to predict the next day’s weather. This dataset contains weather data from January 1, 2000, to December 31, 2009, with eight features, including cloud cover, humidity, sea level pressure, global radiation, rainfall, sunshine, average temperature, and maximum temperature. 

\begin{figure}[h]
 \centering
 \includegraphics[width=0.8\textwidth]{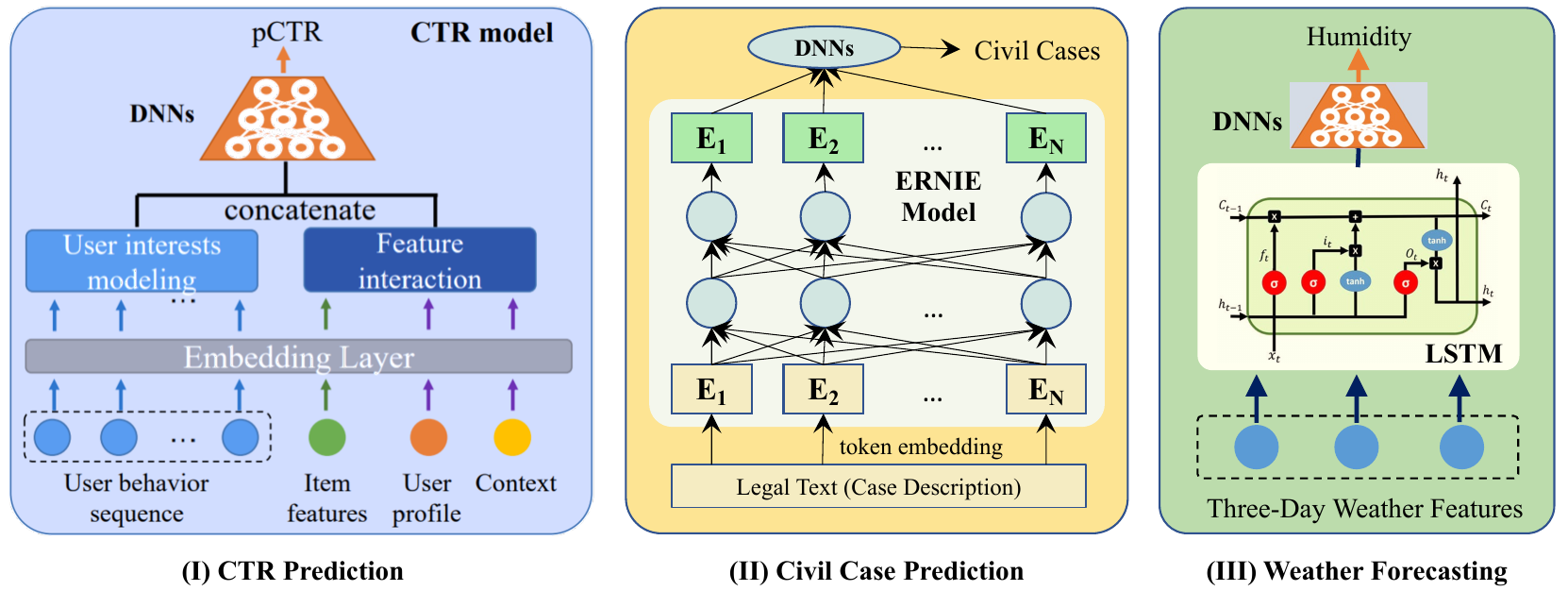}
 \caption{Three Deep Models for Three Tasks: DNNs for CTR Prediction, ERNIE Model for Civil Case Prediction, and Time-Series Model for Weather Forecasting.}
 \label{fig:method2}
\end{figure}

As shown in Figure \ref{fig:method2}, we use three series of deep models for these tasks: DNNs for CTR Prediction, ERNIE Model for Civil Case Prediction, and Time-Series Model (e.g., CNN-LSTM, GRU, LSTM, and Autoformer) for Weather Forecasting.

\subsection{General Information Metrics on Three Scenario Tasks}

Although OIT provides the mathematical definitions for 11 general information metrics \parencite{xu2022foundations,xu2023objective,xu2024research}(Table \ref{tab:OIT}), these definitions are expressed using abstract mathematical concepts to ensure the broadest adaptability. In practical applications, it is necessary to integrate domain knowledge and the specific characteristics of actual data to clarify the calculation methods for each type of metric, facilitating the evaluation and selection of training data. By mapping the meanings of each scenario's dataset to the 11 metric calculation formulas defined by OIT, we derived the methods for calculating various metrics for any dataset in the three experimental scenarios, as presented in Table \ref{tab:GIME}.

\begin{table}[h]
 \centering
 \caption{\centering 11 General Information Metrics on Three Scenario Tasks}
 \label{tab:GIME}
 \begin{tabular}{c}
\includegraphics[width=0.85\textwidth]{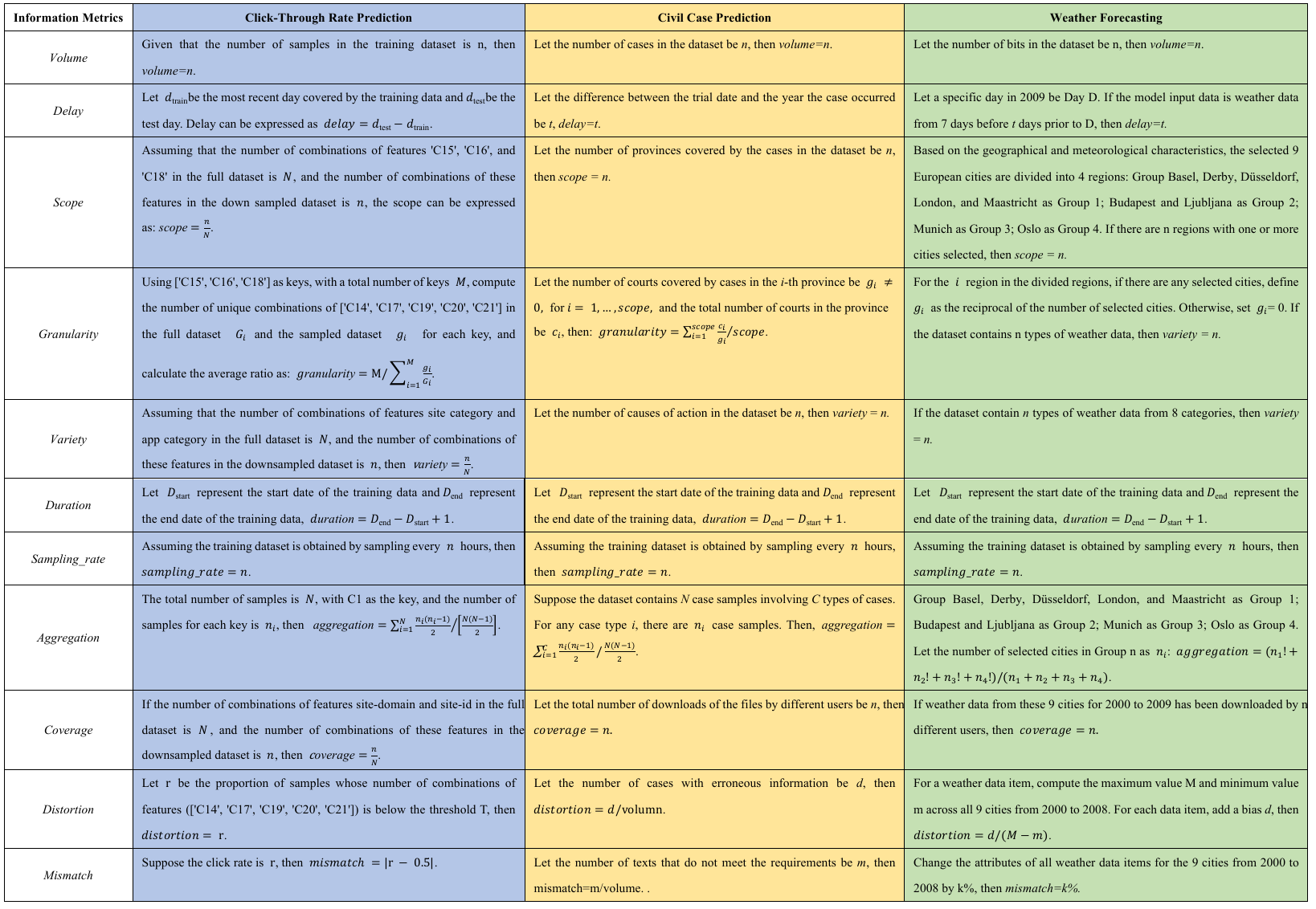} \\
\end{tabular}
\end{table}

\subsection{Exploring the relationship between data metrics and model performance}

\begin{figure}[t]
 \centering \includegraphics[width=\textwidth]{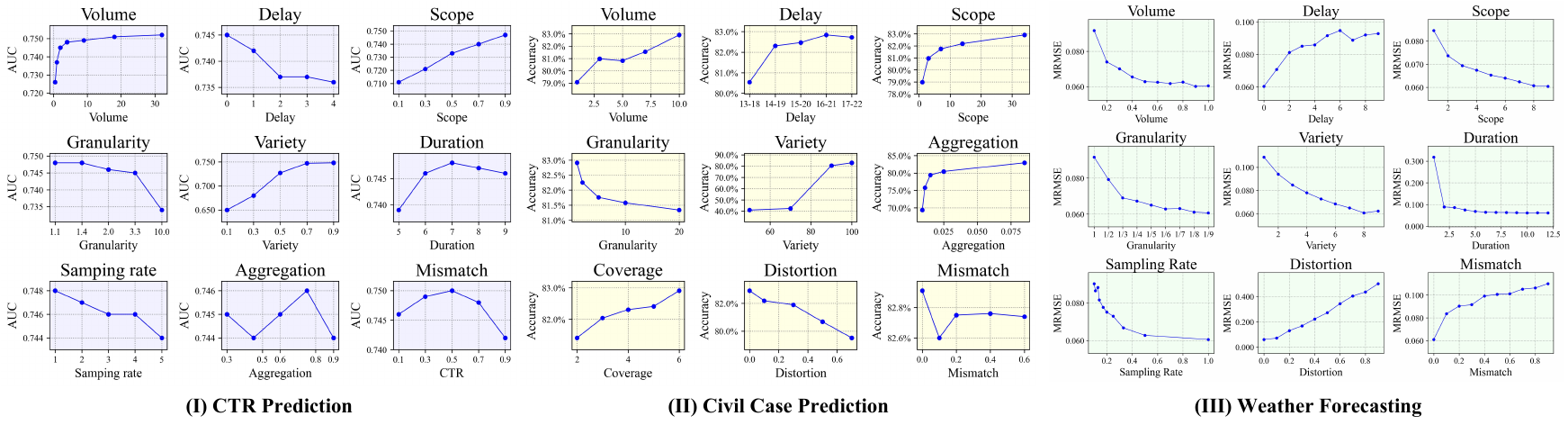}
 \caption{Correlation Experiments Between Training Dataset Information Metrics and AI Model Performance across Three Tasks. (I) CTR Prediction Experiment: Correlation experiments measured AUC against nine metrics, i.e., \textit{volume}, \textit{delay}, \textit{scope}, \textit{granularity}, \textit{variety}, \textit{duration}, \textit{sampling rate}, \textit{aggregation}, and \textit{mismatch}. Note, in the \textit{mismatch} experiments, the horizontal axis represents the training data’s CTR, as \textit{mismatch} is defined as |CTR - 0.5| for intuitive visualization.
 (II) Civil Case Prediction Experiment: Correlation experiments measured Accuracy against nine metrics, i.e., \textit{volume}, \textit{delay}, \textit{scope}, \textit{granularity}, \textit{variety}, \textit{aggregation}, \textit{coverage}, \textit{distortion}, and \textit{mismatch}. (III) Weather Forecasting Experiment: Correlation experiments measured MRMSE against nine metrics, i.e., \textit{volume}, \textit{delay}, \textit{scope}, \textit{granularity}, \textit{variety}, \textit{duration}, \textit{sampling rate}, \textit{distortion}, and \textit{mismatch}. }
 \label{fig:corr}
\end{figure}

As illustrated in Figure \ref{fig:corr}, we utilized DNN~\parencite{covington2016deep}, ERNIE~\parencite{zhang2019ernie}, and LSTM~\parencite{hochreiter1997long} models to evaluate performance in each of these tasks, providing insights into how different training data metrics influence the effectiveness of these predictive models. Since the metric \textit{volume} is correlated with \textit{scope}, \textit{granularity}, \textit{variety}, \textit{duration}, and \textit{sampling rate}, its variation in experiments is achieved by adjusting the aforementioned related metrics. As the other 10 metrics are mutually independent, experiments on the impact of their individual variations on model performance are conducted by fixing the remaining metrics, ensuring objective results that are mutually independent.

 In CTR Prediction, as shown in Figure \ref{fig:corr}(I), we observed a strong correlation between model performance (AUC) and seven key metrics, including \textit{volume}, \textit{delay}, \textit{scope}, \textit{granularity}, \textit{variety}, \textit{sampling} \textit{rate}, and \textit{mismatch}. Monotonic changes in these metrics consistently impacted the model’s performance. In contrast, an increase in \textit{duration} leads to an initial rise in AUC followed by a decline. This is likely because some earlier behaviors no longer match the characteristics of current behaviors, and using such data for training reduces the model’s accuracy in predicting current behaviors. On the other hand, the increase in \textit{aggregation} causes only random fluctuations in AUC, as this metric is not correlated with model performance. 
 We did not evaluate the \textit{distortion} and \textit{coverage} metrics, although these metrics are known to affect model performance. However, due to the lack of error information in the Avazu dataset, it was challenging to calculate the \textit{distortion} metric. Additionally, we did not conduct experiments on \textit{coverage}, as it is difficult to access and is unlikely to have a significant impact on model performance in this context. For the same reasons, the \textit{coverage} metric was also not considered in the subsequent two experiments.
 
 Using a dataset of 100,000 legal documents sourced from China Judgments Online, we investigated how model accuracy in Civil Case Prediction is influenced by nine data metrics. As shown in Figure \ref{fig:corr}(II), monotonic changes in eight metrics—\textit{volume}, \textit{delay}, \textit{scope}, \textit{granularity}, \textit{variety}, \textit{duration}, \textit{aggregation}, and \textit{distortion}—were strongly correlated with model accuracy. 
 However, increasing \textit{mismatch} leads to random fluctuations in model accuracy, with all such fluctuations remaining within a range of 0.4\%. This is because as long as \textit{mismatch} stays within a reasonable range, its variations do not cause significant differences in model performance.
 We did not test for \textit{duration} and \textit{sampling rate} as these temporal metrics are not relevant for Civil Case Predictions due to the long-term stability of legal rulings.
 
 Using weather data from nine European cities between 2000 and 2008, we predicted eight weather variables for future days, evaluating model performance using the Mean Root Mean Square Error (MRMSE). Similarly, we selected only nine metrics for experimentation. As shown in Figure 2(c), monotonic changes in nine metrics—\textit{volume}, \textit{delay}, \textit{granularity}, \textit{variety}, \textit{duration}, \textit{sampling} \textit{rate}, \textit{distortion}, and \textit{mismatch}—were closely correlated with model performance (MRMSE). For the same reasons as in the CTR Prediction experiment, the \textit{coverage} metric was not considered due to the difficulty of obtaining \textit{coverage} indicators for the weather data from the nine cities, and because this metric does not impact the performance of the forecasting model. Additionally, since the \textit{aggregation} metric reflects the geographic distribution characteristics of the nine cities and has a strong correlation with granularity, it was also excluded from our experiments.

While CTR Prediction, Civil Case Prediction, and Weather Forecasting represent fundamentally different domains, our experiments investigating the correlation between data metrics and model performance reveal a unifying pattern. Across all three fields, at least seven of the 11 OIT-defined metrics exhibited monotonic relationships with model performance. These results suggest that the 11 general data metrics could provide a versatile framework for assessing the quality of training datasets across a range of AI models and applications.

\subsection{Using GIME for Improving Model Training Efficiency}

The predominant approach for training AI models relies on using the entire dataset~\parencite{gebru2021datasheets,motamedi2021data}. The GIME method, however, seeks to systematically enhance training efficiency by integrating theoretical principles and practical guidelines. At its core, GIME leverages the 11 general information metrics defined by OIT, calculating these metrics for the training dataset in alignment with domain-specific knowledge. Metrics are then categorized by sensitivity level—high, moderate, or low. For high-sensitivity metrics, optimization is prioritized; moderate-sensitivity metrics are managed by setting reasonable thresholds, while low-sensitivity metrics are excluded. Training proceeds only once all sensitive metrics fall within acceptable thresholds, ensuring robust model performance while significantly reducing data size and training time compared to traditional full-dataset approaches.

\begin{figure}[H]
 \centering
 \includegraphics[width=0.8\textwidth]{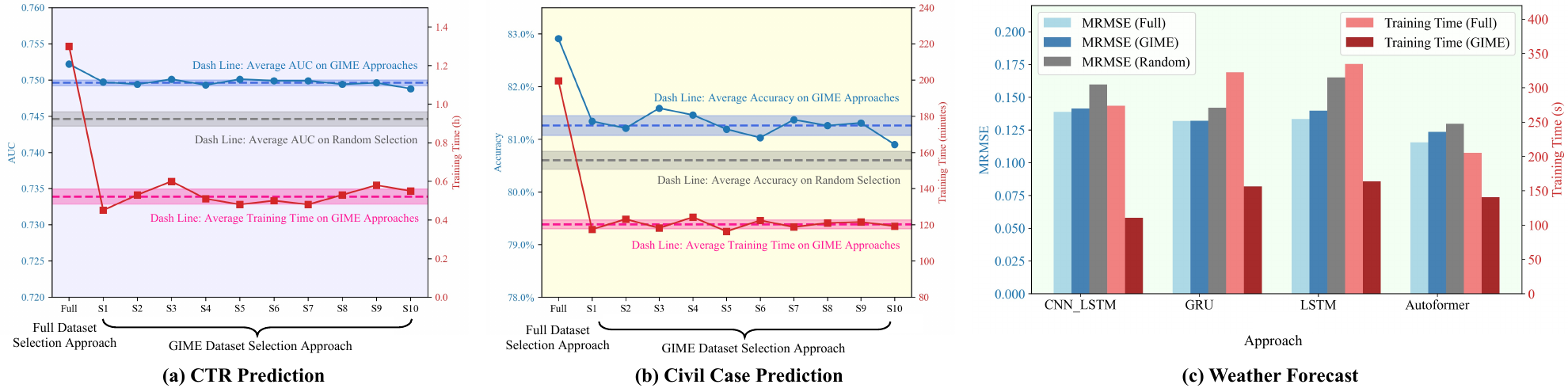}
 \caption{Performance Comparison using GIME-based Dataset Selection across Three Distinct Tasks. (a) \textbf{CTR Prediction}: “Full” indicates models trained on the entire dataset, while S1–S10 denote GIME-selected subsets. (b) \textbf{Civil Case Prediction}: “Full” denotes full dataset training; S1–S10 are GIME subsets. (c) \textbf{Weather Forecasting}: Results for “Full,” GIME, and random sampling across CNN, LSTM, GRU, and Autoformer. In the CTR Prediction and Civil Case Prediction tasks, the dashed lines represent the average results, while the shaded areas denote the standard deviation.}
 \label{exp:GIME}
\end{figure}

For the CTR Prediction task, we identified \textit{delay}, \textit{granularity}, \textit{variety}, and \textit{distortion} as high-sensitivity metrics. These were set to optimal values based on the full data pool, meaning no degradation in performance for these metrics. \textit{Volume}, with a limited impact on performance after a certain threshold, was treated as a moderate-sensitivity metric, with a threshold range between 0.25 and 0.75 of the full data ratio~\parencite{song2020towards,biccici2023efficiently}. Metrics like \textit{duration}, \textit{sampling rate}, \textit{aggregation}, \textit{coverage}, and \textit{mismatch} were classified as low sensitivity and were not considered in data selection. Following these guidelines, one experiment was conducted using the full dataset, and ten experiments were performed with the GIME method. As shown in Figure \ref{exp:GIME}(a), the model trained on the full dataset achieved 0.7522 AUC with a training time of 1.3 hours. The worst-performing model using GIME achieved 0.7488 AUC, a decrease of only 0.0045, while the longest training time was 0.60 hours—reducing training time by 53.8\%. This demonstrates that the GIME method can significantly improve training efficiency with minimal impact on model performance. To further evaluate the effectiveness of the GIME method, we conducted 10 repeated experiments under the same data size, comparing it with the random sampling method. The random method achieved an average AUC of 0.7446 with a standard deviation of 0.001, while the GIME method achieved an average AUC of 0.7496 with a standard deviation of 0.0004. The results demonstrate that GIME consistently outperforms random sampling, with higher AUC and lower variance.

In Civil Case Prediction experiments, the training data pool metrics were as follows: \textit{volume} = 100,000, \textit{delay} = 0, \textit{scope} = 32, \textit{granularity} = 1, \textit{variety} = 100, \textit{duration} = 6, \textit{sampling} \textit{rate} = 1, \textit{aggregation} = 0.245, \textit{distortion} = 0, and \textit{mismatch} = 0. Given the critical importance of variety, balance, accuracy, and adaptability for Civil Case Prediction, we identified \textit{variety}, \textit{aggregation}, \textit{distortion}, and \textit{mismatch} as high-sensitivity metrics, requiring them to match the values of the data pool. \textit{Scope}, while not crucial for performance, was deemed a moderate-sensitivity metric with a threshold set at 27 to mitigate potential cross-regional analysis bias. \textit{Volume}, though impactful, was classified as a moderate-sensitivity metric, with a threshold of 60,000, or 60\% of the data pool, to balance data size with model performance. \textit{Granularity}, \textit{duration}, \textit{sampling} \textit{rate}, and \textit{coverage} were considered low-sensitivity metrics and not prioritized. Following these guidelines, we compared GIME with full data selection. Results in Figure \ref{exp:GIME}(b) show the model trained on the full dataset achieved 82.9\% accuracy in 199.52 minutes. In contrast, 10 trials using the GIME method yielded accuracies ranging from 80.90\% to 81.59\%, with an average accuracy of 81.26\%. The performance drop compared to the full data selection method ranged from 1.58\% to 2.41\%, with an average of 1.98\%. The training time for GIME trials ranged from 116.33 to 124.23 minutes, with an average of 120.24 minutes, resulting in a time saving of 37.73\% to 41.69\%, with an average of 39.73\%. We also compared the GIME method with the random sampling method using the same data size over 10 repeated trials. The random method achieved an average accuracy of 80.60\% with a standard deviation of 0.17\%, while the GIME method achieved an average accuracy of 81.26\% with a standard deviation of 0.19\%. These results show that GIME consistently outperforms the random method in terms of stability, with definitely more competitive accuracy.

In the Weather Forecasting experiment, based on the characteristics of weather data and the nine cities included, we identified \textit{scope}, \textit{granularity}, \textit{variety}, \textit{delay}, \textit{distortion}, and \textit{mismatch} as high-sensitivity metrics, as their variations strongly affect model training. \textit{Volume}, \textit{duration}, and \textit{sampling} \textit{rate} were treated as moderate-sensitivity metrics, with \textit{volume} thresholds set between 40\% and 70\% of the total data pool. \textit{Aggregation} and \textit{coverage} were considered low-sensitivity metrics. The data pool comprised weather data from 2000–2008 across nine European cities, with high-sensitivity metrics calculated as \textit{delay} = 0, \textit{scope} = 4, \textit{granularity} = 0.6458, \textit{variety} = 8, \textit{distortion} = 0, and \textit{mismatch } = 0. We applied the GIME method with CNN\_LSTM, GRU, LSTM, and Transformer-based Autoformer models. As shown in Figure \ref{exp:GIME}(c), full dataset training error ranged from 11.56\% to 13.88\%, with Autoformer performing best. GIME-adjusted models saw a slight performance decrease (0.17\%–6.91\%) but training time was reduced by 31.59\%–59.60\%, significantly improving efficiency. The GIME method outperformed the random sampling method across all tested models, including CNN\_LSTM, GRU, LSTM, and Autoformer. Specifically, GIME reduced the mean RMSE (MRMSE) by 0.0184, 0.0101, 0.0254, and 0.0061 for CNN\_LSTM, GRU, LSTM, and Autoformer, respectively, compared to random sampling. This demonstrates that GIME not only improves prediction accuracy by lowering error rates but also achieves consistent performance enhancements across diverse model architectures, reinforcing its robustness and adaptability to various forecasting tasks.

\subsection{Comparing Active Learning Method for AI Model Training Data Selection}

The modAL algorithm is a popular active learning framework, designed with modularity, flexibility, and extensibility. Built on top of scikit-learn, it enables researchers to quickly create active learning workflows with a high degree of freedom~\parencite{danka2018modal}. 
In this paper, we choose modAL as a baseline model to compare its performance with our GIME. Additionally, researchers can easily replace components with custom-built solutions, facilitating the development of novel algorithms. The modAL method starts by collecting a large dataset. The next key step is selecting an appropriate model. After the first round of selected data, also known as labeled data, supports model training, the accuracy is assessed. If the model’s accuracy does not meet the required threshold, new data is selected based on uncertainty criteria, added to the training labels, and the process is repeated until the model’s accuracy satisfies the set requirements.

\begin{figure}[H]
 \centering
 \includegraphics[width=0.8\textwidth]{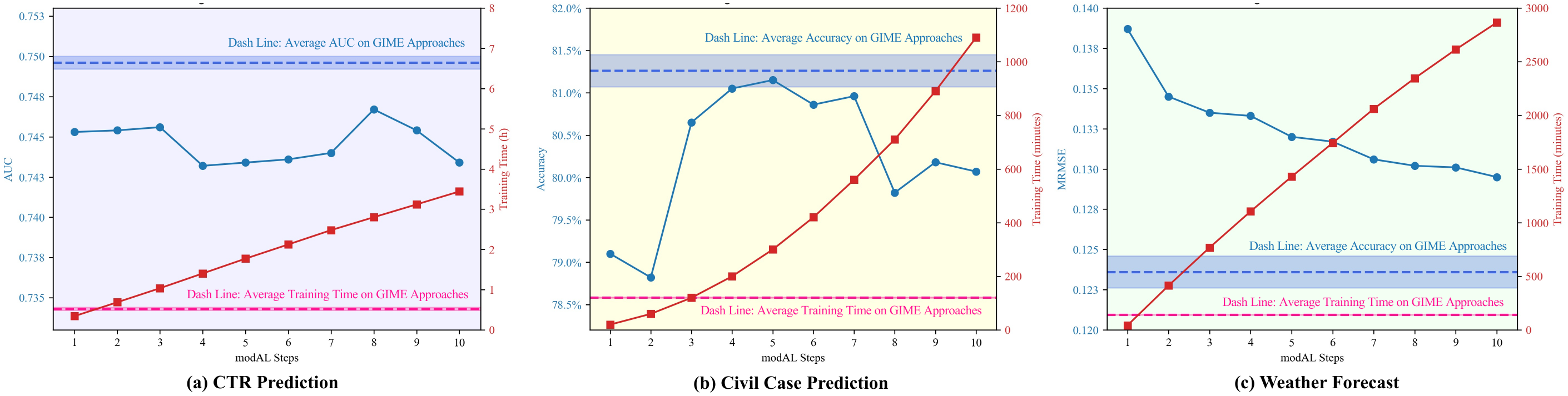}
 \caption{Experimental Comparison of GIME,  modAL Methods across Three Tasks.
 (a) CTR Prediction Results across modAL Steps. (b) Civil Case Prediction Results across modAL Steps. (c) Weather Forecasting Results across modAL Steps. In these three tasks, the dashed lines represent the average results, while the shaded areas denote the standard deviation of each metric.}
 \label{ex:modAL}
\end{figure}

As shown in Figure \ref{ex:modAL}(a), in the CTR Prediction task, GIME demonstrated a clear advantage over modAL in terms of accuracy, stability, and training efficiency. GIME achieved an average AUC of 0.7496 with a remarkably low standard deviation of 0.0004, indicating both high accuracy and exceptional consistency across trials. In contrast, modAL’s performance was less consistent, with its AUC values fluctuating within the range of 0.7433–0.7467 over 10 trials and an average AUC of 0.7444. Although modAL reached its highest AUC of 0.7467 in trial 7, it was still outperformed by GIME, and the larger variability in modAL’s results suggests greater sensitivity to initial conditions. In the Civil Case Prediction experiment, we initially selected a random sample of 10,000 cases to train the ERNIE model, using modAL’s entropy-based selection to iteratively add data in 10,000-sample batches until the full dataset was utilized (see Figure \ref{ex:modAL}(b)). From the results, GIME demonstrated clear superiority in both accuracy and stability compared to modAL. The average accuracy achieved by GIME across 10 trials was 81.26\%, with a relatively narrow range (80.90\%–81.59\%), indicating consistent and reliable performance. In contrast, modAL achieved an average accuracy of 80.26\%, with a wider range (78.82\%–81.15\%), highlighting greater variability and less stability in its results. While modAL reached a peak accuracy of 81.15, close to GIME’s average performance, the modAL method requires at least 4 steps, totaling 200 minutes—80 minutes more than the GIME approach. In comparison, the GIME method reduces the time cost by 40\% while maintaining the same performance. For Weather Forecasting, modAL employed uncertainty sampling, selecting 3,000 samples per iteration to train the Autoformer model (see Figure \ref{ex:modAL}(c)). The best performance achieved by modAL, as measured by MRMSE, was 0.1295, while the error upper bound of GIME was 0.1246, demonstrating that modAL consistently underperformed compared to GIME on the Weather Forecasting dataset. The results underscore GIME’s superiority in handling this task, achieving better error minimization and greater reliability, further reflecting its advanced optimization capabilities and robustness across diverse datasets and tasks.

\subsection{Application of the GIME Method in Judicial AI Program}

\begin{figure}[H]
 \centering
 \includegraphics[width=0.7\textwidth]{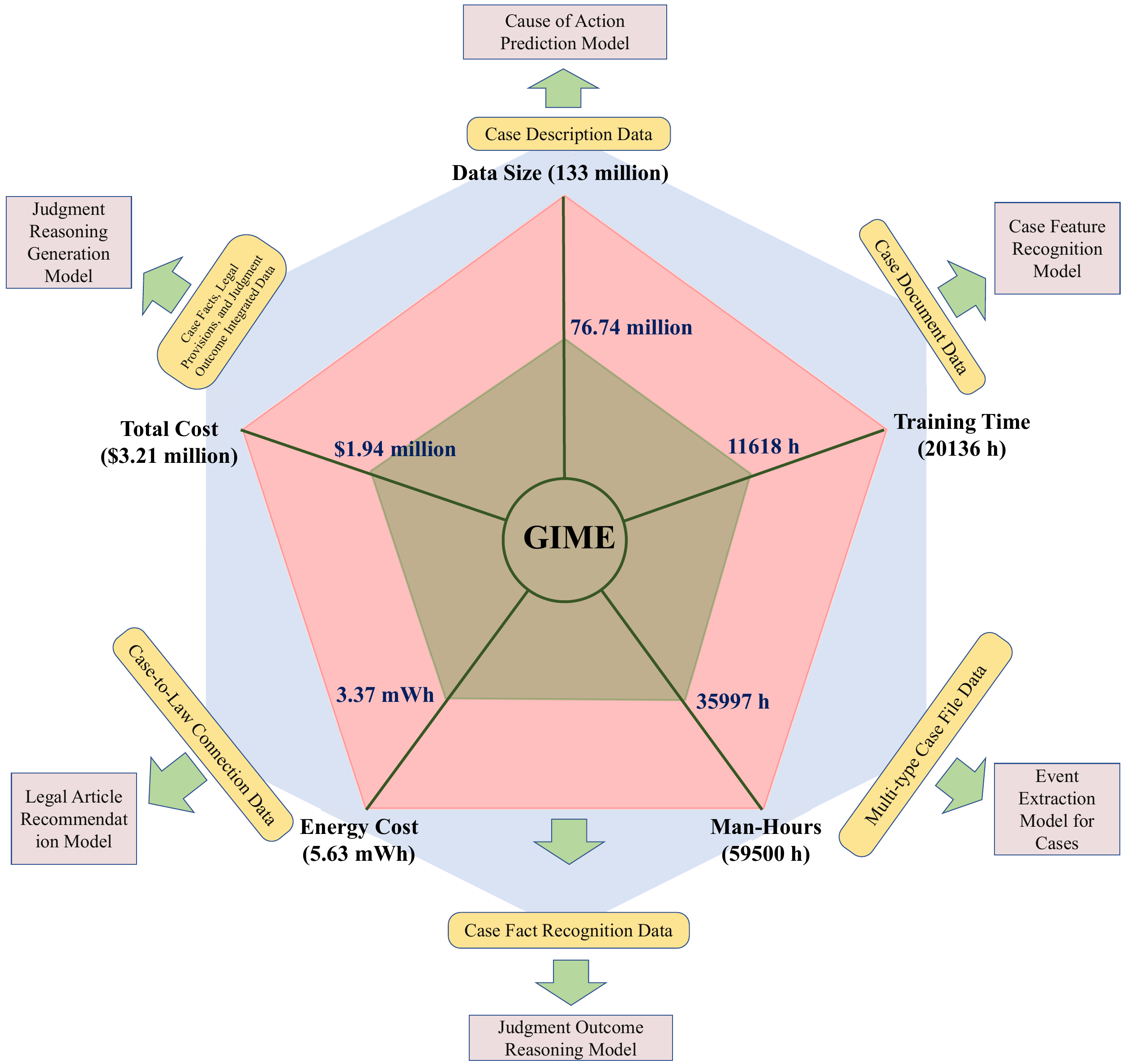}
 \caption{Application and Effectiveness of the GIME Method in Judicial AI Program Development. The GIME method was applied to support the evaluation and selection of six types of training data: Case Description Data, Case Document Data, Multi-type Case File Data, Case Fact Recognition Data, Case-to-Law Connection Data, and Case Facts, Legal Provisions, and Judgment Outcome Integrated Data. These datasets were used to train six models, respectively: the Cause of Action Prediction Model, Case Feature Recognition Model, Event Extraction Model for Cases, Judgment Outcome Reasoning Model, Legal Article Recommendation Model, and Judgment Reasoning Generation Model. In the accompanying visual representation, the pink pentagon’s five vertices correspond to the estimated values for the following resources when using the full dataset selection method: data size, training time, man-hours, energy consumption, and total cost. The purple pentagon’s five vertices represent the values statistically obtained using the GIME method, reflecting reduced resource consumption across five aspects during the model training process.}
 \label{mod:real}
\end{figure}

Since 2017, the China Judicial Big Data Research Institute has led the development and ongoing refinement of six critical AI models tailored to various judicial applications. This initiative, carried out by collaborative teams of technical experts and legal professionals, integrates the GIME method with domain-specific insights and advanced AI technologies. The six AI models—designed for civil case analysis, case feature recognition, event extraction, judgment reasoning, relevant legal provision recommendation, and judgment generation—have been successfully deployed across multiple judicial AI systems. Each model was trained on distinct datasets crafted for its specific function, including case descriptions, judicial documents, multi-type case files, fact-finding data, case-law correlation data, and integrated datasets that combine case facts, legal provisions, and judicial outcomes (see Figure \ref{mod:real}). Based on comprehensive analysis, we estimate that the GIME method, compared to the previously standard full-data selection approach, has reduced training data size by approximately 56.26 million records. This reduction has translated into significant efficiency gains, including a decrease of 8,518 training hours, a savings of 23,503 human hours, and a reduction in energy consumption by 2.26 million watt-hours. Additionally, adopting the GIME method has contributed to an estimated cost savings of \$1.27 million in research and development.

\section{Discussions}
In selecting training data for AI models, attributes such as completeness, heterogeneity, variety, accuracy, timeliness, and balance are commonly considered. However, the exact definitions, mathematical formulations, and applicability of these attributes remain subjects of debate. There is no consensus on whether these characteristics are sufficient, which specific contexts they best serve, or whether a universal method could guide data selection across industries to enhance AI training efficiency. As AI adoption continues to grow, addressing inefficiencies in training time and energy consumption has become increasingly urgent.

Data serves as a fundamental representation of information, and effectively addressing these challenges requires establishing a comprehensive information measurement framework for AI model training. Shannon’s information theory introduced entropy as a measure, which has been applied in data selection. However, entropy alone—or even in its modified forms, such as relative entropy or conditional entropy—depends on prior knowledge of the data’s probability distribution, a condition that is often challenging to meet. Consequently, entropy proves insufficient for selecting training data within the diverse AI landscape. Our research employs the 11 general information metrics defined by OIT across three distinct domains—CTR Prediction, Civil Case Prediction, and Weather Forecasting. These metrics provide a robust, quantifiable foundation for assessing training datasets. In all cases, we observed a monotonic relationship between model performance and most data metrics, confirming that model outcomes are influenced predictably by certain training data characteristics. This principle demonstrates broad applicability across various fields and AI models.

Building on these findings, we proposed the GIME method as a model-agnostic approach to enhance AI training data selection across various domains. Based on common sense and probability theory, we have demonstrated that the GIME method offers advantages over the commonly used full-data selection and random sampling methods by reducing training time and improving model performance, respectively. In our experiments with six models, including Transformer-based architectures, GIME consistently reduced data size and training time while maintaining near-optimal model performance.Comparison experiments across the three scenarios with the random sampling method also demonstrate that, given the same training time, the model performance achieved by the GIME method is significantly superior. This validates the correctness of the theoretical analysis. We further validated GIME by comparing it with the modAL active learning framework in the three scenario tasks. The results also demonstrated significant time savings with GIME while achieving comparable performance levels. We further validated GIME by comparing it with the modAL active learning framework in Civil Case Prediction and Weather Forecasting tasks. These comparisons demonstrated significant time savings with GIME while achieving comparable performance levels. 

GIME also proves its practicality in large-scale engineering applications, particularly in the field of judicial AI in China. Here, GIME was integrated into the development of six critical AI models for tasks such as civil case analysis, case feature recognition, event extraction, judgment reasoning, relevant legal provision recommendation, and judgment generation—have been successfully deployed across multiple judicial AI systems. By systematically evaluating and optimizing training datasets, GIME reduced data size by 56.26 million records, training hours by 8,518, and energy consumption by 2.26 million watt-hours. These efficiency gains translated into significant cost reductions, saving an estimated \$1.27 million in research and development expenses. The success of these applications underscores GIME’s potential as a transformative tool for resource-efficient AI development, particularly in domains where computational costs and timelines are critical considerations.

While GIME introduces a pre-training evaluation phase, as the dataset metrics are calculated using simple mathematical formulas, its computational overhead is relatively minor compared to the savings achieved during the training process. The additional step of evaluating data metrics is offset by the reduced need for redundant computations and the smaller training datasets required. For example, in Weather Forecasting, GIME-adjusted models achieved training time reductions of up to 59.6\%, significantly outweighing the computational cost of the initial evaluation. Furthermore, the modularity of GIME allows for the selective prioritization of high-sensitivity metrics, ensuring that the evaluation phase remains efficient even for large-scale datasets. Overall, the benefits of improved training efficiency, reduced resource consumption, and enhanced predictability far outweigh the modest computational costs incurred.

\section{Conclusion}

This study presents a systematic approach to optimizing AI training data selection through the 11 general information metrics defined by OIT and the GIME method. These metrics provide a quantifiable framework for assessing data quality, while GIME offers a practical implementation for diverse AI applications. Experimental results across three representative tasks such as Human Behavior tasks, Natural Language Understanding tasks, and Time-Series tasks consistently demonstrated GIME’s ability to reduce data size, training time, and energy consumption in large-scale complex tasks while maintaining near-optimal performance. The successful deployment of GIME in engineering applications, particularly judicial AI systems, highlights its real-world impact, including significant reductions in costs, resource usage, and development timelines.

OIT offers a new perspective to define the concept of information, forming the most comprehensive and systematic universal information metrics. This paper is the first to fully use these metrics to support the evaluation of AI training data, showing that the GIME method can significantly improve training efficiency across different fields and models. Therefore, the significance of this work extends beyond AI technology, inspiring academic peers to further understand and examine the intrinsic nature and measurement methods of information. It also lays the groundwork for deeper exploration and characterization of the dynamic mechanisms of information systems, providing new theories, methods, and pathways for achieving more groundbreaking advancements in information technology, exemplified by AI.

Despite its demonstrated benefits, GIME’s reliance on domain-specific knowledge for defining threshold values may limit its applicability in domains where such expertise is unavailable. Additionally, the method has primarily been evaluated in static training environments, leaving room for further exploration in dynamic or adaptive AI systems. Future research should focus on automating the threshold-setting process, potentially through machine learning techniques, to enhance GIME’s generalizability. Expanding the application of GIME to more complex and heterogeneous datasets, as well as integrating it with real-time learning systems, could further extend its utility. By addressing these limitations, GIME has the potential to evolve into a cornerstone methodology for sustainable and efficient AI development on a global scale.

\printbibliography

@incollection{bialkova2024ai,
  title={AI transforming business and everyday life},
  author={Bialkova, Svetlana},
  booktitle={The rise of AI user applications: Chatbots integration foundations and trends},
  pages={143--165},
  year={2024},
  publisher={Springer},
  doi={https://doi.org/10.1007/978-3-031-56471-0_9}
}

@article{gunasekar2023textbooksneed,
      title={Textbooks Are All You Need}, 
      author={Gunasekar, Suriya and Zhang, Yi and Aneja, Jyoti and Mendes, Caio C{\'e}sar Teodoro and Del Giorno, Allie and Gopi, Sivakanth and Javaheripi, Mojan and Kauffmann, Piero and de Rosa, Gustavo and Saarikivi, Olli and others},
      journal={arXiv preprint arXiv:2306.11644},
      year={2023},
      doi={https://doi.org/10.48550/arXiv.2306.11644}
}

@article{li2023textbooksneediiphi15,
      title={Textbooks Are All You Need II: phi-1.5 technical report}, 
      author={Li, Yuanzhi and Bubeck, S{\'e}bastien and Eldan, Ronen and Giorno, AD and Gunasekar, Suriya and Lee, Yin Tat},
      journal={arXiv preprint arXiv:2309.05463},
      volume={5},
      year={2023},
      doi={https://doi.org/10.48550/arXiv.2309.05463}
}

@article{pouyanfar2018survey,
  title={A survey on deep learning: Algorithms, techniques, and applications},
  author={Pouyanfar, Samira and Sadiq, Saad and Yan, Yilin and Tian, Haiman and Tao, Yudong and Reyes, Maria Presa and Shyu, Mei-Ling and Chen, Shu-Ching and Iyengar, Sundaraja S},
  journal={ACM computing surveys (CSUR)},
  volume={51},
  number={5},
  pages={1--36},
  year={2018},
  publisher={ACM New York, NY, USA},
  doi={https://doi.org/10.1145/3234150}
}

@article{minaee2024large,
  title={Large language models: A survey},
  author={Minaee, Shervin and Mikolov, Tomas and Nikzad, Narjes and Chenaghlu, Meysam and Socher, Richard and Amatriain, Xavier and Gao, Jianfeng},
  journal={arXiv preprint arXiv:2402.06196},
  year={2024},
  doi={https://doi.org/10.48550/arXiv.2402.06196}
}

@article{touvron2023llama,
  title={Llama: Open and efficient foundation language models},
  author={Touvron, Hugo and Lavril, Thibaut and Izacard, Gautier and Martinet, Xavier and Lachaux, Marie-Anne and Lacroix, Timoth{\'e}e and Rozi{\`e}re, Baptiste and Goyal, Naman and Hambro, Eric and Azhar, Faisal and others},
  journal={arXiv preprint arXiv:2302.13971},
  year={2023},
  doi={https://doi.org/10.48550/arXiv.2302.13971}
}

@article{touvron2023llama2,
  title={Llama 2: Open foundation and fine-tuned chat models},
  author={Touvron, Hugo and Martin, Louis and Stone, Kevin and Albert, Peter and Almahairi, Amjad and Babaei, Yasmine and Bashlykov, Nikolay and Batra, Soumya and Bhargava, Prajjwal and Bhosale, Shruti and others},
  journal={arXiv preprint arXiv:2307.09288},
  year={2023},
  doi={https://doi.org/10.48550/arXiv.2307.09288}
}

@article{chowdhery2023palm,
  title={Palm: Scaling language modeling with pathways},
  author={Chowdhery, Aakanksha and Narang, Sharan and Devlin, Jacob and Bosma, Maarten and Mishra, Gaurav and Roberts, Adam and Barham, Paul and Chung, Hyung Won and Sutton, Charles and Gehrmann, Sebastian and others},
  journal={Journal of Machine Learning Research},
  volume={24},
  number={240},
  pages={1--113},
  year={2023}
}

@article{shorten2019survey,
  title={A survey on image data augmentation for deep learning},
  author={Shorten, Connor and Khoshgoftaar, Taghi M},
  journal={Journal of big data},
  volume={6},
  number={1},
  pages={1--48},
  year={2019},
  publisher={Springer},
  doi={https://doi.org/10.1186/s40537-019-0197-0}
}

@article{jakubik2024data,
  title={Data-centric artificial intelligence},
  author={Jakubik, Johannes and V{\"o}ssing, Michael and K{\"u}hl, Niklas and Walk, Jannis and Satzger, Gerhard},
  journal={Business \& Information Systems Engineering},
  pages={1--9},
  year={2024},
  publisher={Springer},
  doi={https://doi.org/10.1007/s12599-024-00857-8}
}

@article{danka2018modal,
  title={modAL: A modular active learning framework for Python},
  author={Danka, Tivadar and Horvath, Peter},
  journal={arXiv preprint arXiv:1805.00979},
  year={2018},
  doi={https://doi.org/10.48550/arXiv.1805.00979}
}

@article{zhan2022comparative,
  title={A comparative survey of deep active learning},
  author={Zhan, Xueying and Wang, Qingzhong and Huang, Kuan-hao and Xiong, Haoyi and Dou, Dejing and Chan, Antoni B},
  journal={arXiv preprint arXiv:2203.13450},
  year={2022},
  doi={https://doi.org/10.48550/arXiv.2203.13450}
}

@book{xu2023objective,
  title={Objective Information Theory},
  author={Xu, Jianfeng and Wang, Shuliang and Liu, Zhenyu and Wang, Yashi and Wang, Yingfei and Dang, Yingxu},
  year={2023},
  publisher={Springer Nature}
}

@book{ash2012information,
  title={Information theory},
  author={Ash, Robert B},
  year={2012},
  publisher={Courier Corporation}
}

@article{singh2023systematic,
  title={Systematic review of data-centric approaches in artificial intelligence and machine learning},
  author={Singh, Prerna},
  journal={Data Science and Management},
  volume={6},
  number={3},
  pages={144--157},
  year={2023},
  publisher={Elsevier},
  doi={https://doi.org/10.1016/j.dsm.2023.06.001}
}

@article{gunasekar2023textbooks,
  title={Textbooks are all you need},
  author={Gunasekar, Suriya and Zhang, Yi and Aneja, Jyoti and Mendes, Caio C{\'e}sar Teodoro and Del Giorno, Allie and Gopi, Sivakanth and Javaheripi, Mojan and Kauffmann, Piero and de Rosa, Gustavo and Saarikivi, Olli and others},
  journal={arXiv preprint arXiv:2306.11644},
  year={2023},
  doi={https://doi.org/10.48550/arXiv.2306.11644}
}

@article{zha2023data,
  title={Data-centric artificial intelligence: A survey},
  author={Zha, Daochen and Bhat, Zaid Pervaiz and Lai, Kwei-Herng and Yang, Fan and Jiang, Zhimeng and Zhong, Shaochen and Hu, Xia},
  journal={arXiv preprint arXiv:2303.10158},
  year={2023},
  doi={https://doi.org/10.48550/arXiv.2303.10158}
}

@article{hamid2023data,
  title={Data-centric and model-centric AI: Twin drivers of compact and robust industry 4.0 solutions},
  author={Hamid, Oussama H},
  journal={Applied Sciences},
  volume={13},
  number={5},
  pages={2753},
  year={2023},
  publisher={MDPI},
  doi={https://doi.org/10.3390/app13052753}
}

@article{achiam2023gpt,
  title={Gpt-4 technical report},
  author={Achiam, Josh and Adler, Steven and Agarwal, Sandhini and Ahmad, Lama and Akkaya, Ilge and Aleman, Florencia Leoni and Almeida, Diogo and Altenschmidt, Janko and Altman, Sam and Anadkat, Shyamal and others},
  journal={arXiv preprint arXiv:2303.08774},
  year={2023},
  doi={https://doi.org/10.48550/arXiv.2303.08774}
}

@article{zhao2023survey,
  title={A survey of large language models},
  author={Zhao, Wayne Xin and Zhou, Kun and Li, Junyi and Tang, Tianyi and Wang, Xiaolei and Hou, Yupeng and Min, Yingqian and Zhang, Beichen and Zhang, Junjie and Dong, Zican and others},
  journal={arXiv preprint arXiv:2303.18223},
  year={2023},
  doi={https://doi.org/10.48550/arXiv.2303.18223}
}

@article{dong2021survey,
  title={A survey on deep learning and its applications},
  author={Dong, Shi and Wang, Ping and Abbas, Khushnood},
  journal={Computer Science Review},
  volume={40},
  pages={100379},
  year={2021},
  publisher={Elsevier},
  doi={https://doi.org/10.1016/j.cosrev.2021.100379}
}

@inproceedings{song2020towards,
  title={Towards automated neural interaction discovery for click-through rate prediction},
  author={Song, Qingquan and Cheng, Dehua and Zhou, Hanning and Yang, Jiyan and Tian, Yuandong and Hu, Xia},
  booktitle={Proceedings of the 26th ACM SIGKDD International Conference on Knowledge Discovery \& Data Mining},
  pages={945--955},
  year={2020},
  doi={https://doi.org/10.1145/3394486.3403137}
}

@inproceedings{biccici2023efficiently,
  title={Efficiently Sampling in Neural Network Training for Click-Through Rate Prediction},
  author={Bi{\c{c}}ici, Ergun and Dilbaz, Serdarcan},
  booktitle={2023 8th International Conference on Computer Science and Engineering (UBMK)},
  pages={469--472},
  year={2023},
  organization={IEEE},
  doi={https://doi.org/10.1109/UBMK59864.2023.10286811}
}

@inproceedings{xu2014objective,
  title={Objective information theory: A Sextuple model and 9 kinds of metrics},
  author={Xu, Jianfeng and Ma, Xuefeng and Shen, Yanli and Tang, Jun and Xu, Bin and Qiao, Yongjie},
  booktitle={2014 Science and information conference},
  pages={793--802},
  year={2014},
  organization={IEEE},
  doi={https://doi.org/10.1109/SAI.2014.6918277}
}

@article{xu2022foundations,
  title={Foundations and applications of information systems dynamics},
  author={Xu, Jianfeng and Liu, Zhenyu and Wang, Shuliang and Zheng, Tao and Wang, Yashi and Wang, Yingfei and Dang, Yingxu},
  journal={Engineering},
  year={2022},
  publisher={Elsevier},
  doi={https://doi.org/10.1016/j.eng.2022.04.018}
}

@article{xu2024research,
  title={Research and Application of General Information Measures Based on a Unified Model},
  author={Xu, Jianfeng},
  journal={IEEE Transactions on Computers},
  year={2024},
  publisher={IEEE},
  doi={https://doi.ieeecomputersociety.org/10.1109/TC.2024.3349650}
}

@inproceedings{renyi1961measures,
  title={On measures of entropy and information},
  author={R{\'e}nyi, Alfr{\'e}d},
  booktitle={Proceedings of the fourth Berkeley symposium on mathematical statistics and probability, volume 1: contributions to the theory of statistics},
  volume={4},
  pages={547--562},
  year={1961},
  organization={University of California Press}
}

@article{kaplan2020scaling,
  title={Scaling laws for neural language models},
  author={Kaplan, Jared and McCandlish, Sam and Henighan, Tom and Brown, Tom B and Chess, Benjamin and Child, Rewon and Gray, Scott and Radford, Alec and Wu, Jeffrey and Amodei, Dario},
  journal={arXiv preprint arXiv:2001.08361},
  year={2020},
  doi={https://doi.org/10.48550/arXiv.2001.08361}
}

@book{wiener2019cybernetics,
  title={Cybernetics or Control and Communication in the Animal and the Machine},
  author={Wiener, Norbert},
  year={2019},
  publisher={MIT press}
}

@article{bandi2023power,
  title={The power of generative ai: A review of requirements, models, input--output formats, evaluation metrics, and challenges},
  author={Bandi, Ajay and Adapa, Pydi Venkata Satya Ramesh and Kuchi, Yudu Eswar Vinay Pratap Kumar},
  journal={Future Internet},
  volume={15},
  number={8},
  pages={260},
  year={2023},
  publisher={MDPI},
  doi={https://doi.org/10.3390/fi15080260}
}

@article{cui2023survey,
  title={A survey on legal judgment prediction: Datasets, metrics, models and challenges},
  author={Cui, Junyun and Shen, Xiaoyu and Wen, Shaochun},
  journal={IEEE Access},
  year={2023},
  publisher={IEEE},
  doi={https://doi.org/10.1109/ACCESS.2023.3317083}
}

@inproceedings{zhang2019ernie,
  title={ERNIE: Enhanced Language Representation with Informative Entities},
  author={Zhang, Zhengyan and Han, Xu and Liu, Zhiyuan and Jiang, Xin and Sun, Maosong and Liu, Qun},
  booktitle={Proceedings of the 57th Annual Meeting of the Association for Computational Linguistics},
  pages={1441--1451},
  year={2019},
  doi={https://doi.org/10.18653/v1/P19-1139}
}

@article{hochreiter1997long,
  title={Long Short-term Memory},
  author={Hochreiter, S},
  journal={Neural Computation MIT-Press},
  year={1997},
  doi={https://doi.org/10.1162/neco.1997.9.8.1735}
}

@article{motamedi2021data,
  title={A data-centric approach for training deep neural networks with less data},
  author={Motamedi, Mohammad and Sakharnykh, Nikolay and Kaldewey, Tim},
  journal={arXiv preprint arXiv:2110.03613},
  year={2021},
  doi={https://doi.org/10.48550/arXiv.2110.03613}
}

@inproceedings{covington2016deep,
  title={Deep neural networks for youtube recommendations},
  author={Covington, Paul and Adams, Jay and Sargin, Emre},
  booktitle={Proceedings of the 10th ACM conference on recommender systems},
  pages={191--198},
  year={2016},
  doi={https://doi.org/10.1145/2959100.2959190}
}

@inproceedings{shi2021smart,
  title={The Smart Court-A New Pathway to Justice in China?},
  author={Shi, Changqing and Sourdin, Tania and Li, Bin},
  booktitle={IJCA},
  volume={12},
  pages={1},
  year={2021},
  organization={HeinOnline},
  doi={https://doi.org/10.36745/ijca.367}
}

@article{yang2022click,
  title={Click-through rate prediction in online advertising: A literature review},
  author={Yang, Yanwu and Zhai, Panyu},
  journal={Information Processing \& Management},
  volume={59},
  number={2},
  pages={102853},
  year={2022},
  publisher={Elsevier},
  doi={https://doi.org/10.1016/j.ipm.2021.102853}
}

@article{abhishek2012weather,
  title={Weather forecasting model using artificial neural network},
  author={Abhishek, Kumar and Singh, Maheshwari Prasad and Ghosh, Saswata and Anand, Abhishek},
  journal={Procedia Technology},
  volume={4},
  pages={311--318},
  year={2012},
  publisher={Elsevier},
  doi={https://doi.org/10.1016/j.protcy.2012.05.047}
}

@inproceedings{ma2021legal,
  title={Legal judgment prediction with multi-stage case representation learning in the real court setting},
  author={Ma, Luyao and Zhang, Yating and Wang, Tianyi and Liu, Xiaozhong and Ye, Wei and Sun, Changlong and Zhang, Shikun},
  booktitle={Proceedings of the 44th International ACM SIGIR Conference on Research and Development in Information Retrieval},
  pages={993--1002},
  year={2021},
  doi={https://doi.org/10.1145/3404835.3462945}
}

@article{guo2017deepfm,
  title={DeepFM: a factorization-machine based neural network for CTR prediction},
  author={Guo, Huifeng and Tang, Ruiming and Ye, Yunming and Li, Zhenguo and He, Xiuqiang},
  journal={arXiv preprint arXiv:1703.04247},
  year={2017},
  doi={https://doi.org/10.48550/arXiv.1703.04247}
}

@article{gebru2021datasheets,
  title={Datasheets for datasets},
  author={Gebru, Timnit and Morgenstern, Jamie and Vecchione, Briana and Vaughan, Jennifer Wortman and Wallach, Hanna and Iii, Hal Daum{\'e} and Crawford, Kate},
  journal={Communications of the ACM},
  volume={64},
  number={12},
  pages={86--92},
  year={2021},
  publisher={ACM New York, NY, USA},
  doi={https://doi.org/10.1145/3458723}
}

@article{sekeroglu2022comparative,
  title={Comparative evaluation and comprehensive analysis of machine learning models for regression problems},
  author={Sekeroglu, Boran and Ever, Yoney Kirsal and Dimililer, Kamil and Al-Turjman, Fadi},
  journal={Data Intelligence},
  volume={4},
  number={3},
  pages={620--652},
  year={2022},
  publisher={MIT Press One Rogers Street, Cambridge, MA 02142-1209, USA journals-info~…},
  doi={https://doi.org/10.1162/dint_a_00155}
}

@techreport{program1976annual,
  title = {Program projects, annual report volume 2 (1975–1976)},
  author = {{Program on Information Resources Policy}},
  institution = {Computation Laboratory, Harvard University},
  address = {Cambridge, MA, USA},
  year = {1976},
  number = {R-76-2},
  type = {Technical Report}
}

@book{shapiro1999information,
  title={Information rules: A strategic guide to the network economy},
  author={Shapiro, Carl},
  year={1999},
  publisher={Harvard Business School Press}
}

@article{feuerriegel2024generative,
  title={Generative ai},
  author={Feuerriegel, Stefan and Hartmann, Jochen and Janiesch, Christian and Zschech, Patrick},
  journal={Business \& Information Systems Engineering},
  volume={66},
  number={1},
  pages={111--126},
  year={2024},
  publisher={Springer},
  doi={https://doi.org/10.1007/s12599-023-00834-7}
}

@inproceedings{cui2024survey,
  title={A survey on multimodal large language models for autonomous driving},
  author={Cui, Can and Ma, Yunsheng and Cao, Xu and Ye, Wenqian and Zhou, Yang and Liang, Kaizhao and Chen, Jintai and Lu, Juanwu and Yang, Zichong and Liao, Kuei-Da and others},
  booktitle={Proceedings of the IEEE/CVF Winter Conference on Applications of Computer Vision},
  pages={958--979},
  year={2024},
  doi={https://doi.ieeecomputersociety.org/10.1109/WACVW60836.2024.00106}
}

@article{tian2024role,
  title={The role of large language models in medical image processing: a narrative review},
  author={Tian, Dianzhe and Jiang, Shitao and Zhang, Lei and Lu, Xin and Xu, Yiyao},
  journal={Quantitative Imaging in Medicine and Surgery},
  volume={14},
  number={1},
  pages={1108},
  year={2024},
  publisher={AME Publications},
  doi={https://doi.org/10.21037/qims-23-892}
}

@article{liu2024afm3d,
  title={Afm3d: An asynchronous federated meta-learning framework for driver distraction detection},
  author={Liu, Sheng and You, Linlin and Zhu, Rui and Liu, Bing and Liu, Rui and Yu, Han and Yuen, Chau},
  journal={IEEE Transactions on Intelligent Transportation Systems},
  year={2024},
  publisher={IEEE},
  doi={https://doi.org/10.1109/TITS.2024.3357138}
}

@article{huang2024federated,
  title={Federated learning-empowered AI-generated content in wireless networks},
  author={Huang, Xumin and Li, Peichun and Du, Hongyang and Kang, Jiawen and Niyato, Dusit and Kim, Dong In and Wu, Yuan},
  journal={IEEE Network},
  year={2024},
  publisher={IEEE},
  doi={https://doi.org/10.1109/MNET.2024.3353377}
}

@article{alzubaidi2024ssp,
  title={SSP: self-supervised pertaining technique for classification of shoulder implants in x-ray medical images: a broad experimental study},
  author={Alzubaidi, Laith and Fadhel, Mohammed A and Hollman, Freek and Salhi, Asma and Santamaria, Jose and Duan, Ye and Gupta, Ashish and Cutbush, Kenneth and Abbosh, Amin and Gu, Yuantong},
  journal={Artificial Intelligence Review},
  volume={57},
  number={10},
  pages={261},
  year={2024},
  publisher={Springer},
  doi={https://doi.org/10.1007/s10462-024-10878-0}
}

@article{okanovic2023repeated,
  title={Repeated random sampling for minimizing the time-to-accuracy of learning},
  author={Okanovic, Patrik and Waleffe, Roger and Mageirakos, Vasilis and Nikolakakis, Konstantinos E and Karbasi, Amin and Kalogerias, Dionysis and G{\"u}rel, Nezihe Merve and Rekatsinas, Theodoros},
  journal={arXiv preprint arXiv:2305.18424},
  year={2023}
}

@article{wongvorachan2023comparison,
  title={A comparison of undersampling, oversampling, and SMOTE methods for dealing with imbalanced classification in educational data mining},
  author={Wongvorachan, Tarid and He, Surina and Bulut, Okan},
  journal={Information},
  volume={14},
  number={1},
  pages={54},
  year={2023},
  publisher={MDPI}
}

@article{jiao2022hierarchical,
  title={Hierarchical sampling for the visualization of large scale-free graphs},
  author={Jiao, Bo and Lu, Xin and Xia, Jingbo and Gupta, Brij Bhooshan and Bao, Lei and Zhou, Qingshan},
  journal={IEEE Transactions on Visualization and Computer Graphics},
  volume={29},
  number={12},
  pages={5111--5123},
  year={2022},
  publisher={IEEE}
}

@article{wang2019efficient,
  title={Efficient Stratified Sampling Graphing Method for Mass Data},
  author={Wang, Jianjun and Zhao, Yingang and Chen, Jun and Zhang, Suqing and Zhao, Xudong and He, Yufei},
  journal={Data Science Journal},
  volume={18},
  pages={56--56},
  year={2019}
}

@article{wan2022r2ci,
  title={R2CI: Information theoretic-guided feature selection with multiple correlations},
  author={Wan, Jihong and Chen, Hongmei and Li, Tianrui and Huang, Wei and Li, Min and Luo, Chuan},
  journal={Pattern Recognition},
  volume={127},
  pages={108603},
  year={2022},
  publisher={Elsevier}
}

@inproceedings{pawluk2019information,
  title={Information-theoretic feature selection using high-order interactions},
  author={Pawluk, Mateusz and Teisseyre, Pawe{\l} and Mielniczuk, Jan},
  booktitle={Machine Learning, Optimization, and Data Science: 4th International Conference, LOD 2018, Volterra, Italy, September 13-16, 2018, Revised Selected Papers 4},
  pages={51--63},
  year={2019},
  organization={Springer}
}

@inproceedings{cubuk2019autoaugment,
  title={Autoaugment: Learning augmentation strategies from data},
  author={Cubuk, Ekin D and Zoph, Barret and Mane, Dandelion and Vasudevan, Vijay and Le, Quoc V},
  booktitle={Proceedings of the IEEE/CVF conference on computer vision and pattern recognition},
  pages={113--123},
  year={2019}
}

@article{borovykh2017conditional,
  title={Conditional time series forecasting with convolutional neural networks},
  author={Borovykh, Anastasia and Bohte, Sander and Oosterlee, Cornelis W},
  journal={arXiv preprint arXiv:1703.04691},
  year={2017}
}

@inproceedings{yoo2019learning,
  title={Learning loss for active learning},
  author={Yoo, Donggeun and Kweon, In So},
  booktitle={Proceedings of the IEEE/CVF conference on computer vision and pattern recognition},
  pages={93--102},
  year={2019}
}

@article{park2020robust,
  title={Robust expected model change for active learning in regression},
  author={Park, Sung Ho and Kim, Seoung Bum},
  journal={Applied Intelligence},
  volume={50},
  pages={296--313},
  year={2020},
  publisher={Springer}
}

@article{cai2016batch,
  title={Batch mode active learning for regression with expected model change},
  author={Cai, Wenbin and Zhang, Muhan and Zhang, Ya},
  journal={IEEE transactions on neural networks and learning systems},
  volume={28},
  number={7},
  pages={1668--1681},
  year={2016},
  publisher={IEEE}
}

@inproceedings{finn2017model,
  title={Model-agnostic meta-learning for fast adaptation of deep networks},
  author={Finn, Chelsea and Abbeel, Pieter and Levine, Sergey},
  booktitle={International conference on machine learning},
  pages={1126--1135},
  year={2017},
  organization={PMLR}
}

\section*{Acknowledgements}
This work was supported by the National Key R\&D Program under Grant No. 2022YFC3340900. We are deeply grateful to Professor Yaohui Jin and Associate Professor Yanyan Xu from the Institute of Artificial Intelligence, Shanghai Jiao Tong University, for their insightful suggestions on experimental methods. We also extend our sincere thanks to Associate Professor Rui Wang from the Department of Computer Science, Shanghai Jiao Tong University, for his valuable guidance.

\section*{Data Availability}
The Click-Through Rate prediction data is available at: \url{https://www.kaggle.com/c/avazu-ctr-prediction/data}.
The Civil Case Prediction data is available at: \url{https://wenshu.court.gov.cn/}.
The Weather Forecasting data is available at: \url{https://www.kaggle.com/datasets/thedevastator/weather-prediction/data}.

\section*{Code Availability}
The codes that support the findings of this study are available from the corresponding author upon request.

\section*{Author Contributions Statement}

Jianfeng Xu conceived and designed the study, oversaw the research process, and led manuscript preparation and revisions. As the corresponding author, Xu provided overall supervision and ensured the integrity of the research and its presentation. Congcong Liu performed the CTR Prediction experiments and contributed to the refinement of the GIME methodology. Xiaoying Tan conducted Civil Case Prediction experiments and assessed the impacts of judicial AI applications. Xiaojie Zhu developed the experimental framework for Weather Forecasting and conducted result analysis. Anpeng Wu collected and reviewed relevant literature, summarized key findings, and assisted in the preparation of the manuscript. Huan Wan was responsible for model training in the weather forecasting experiments. Weijun Kong coordinated and facilitated the execution of the weather forecasting experiments. Chun Li conducted an in-depth analysis and articulation of the research methodology. Hu Xu interpreted and analyzed experimental outcomes. Kun Kuang enhanced the overall research framework and revised partial experiment designs. Fei Wu critically reviewed and approved the research design and the final manuscript.

\section*{Competing interests} 
The authors declare no competing interests.

\end{document}